\documentclass[10pt,twocolumn,letterpaper]{article}

\usepackage{wacv}
\usepackage{times}
\usepackage{epsfig}
\usepackage{graphicx}
\usepackage{amsmath}
\usepackage{amssymb}

\usepackage{fancyhdr}

\usepackage{booktabs}
\usepackage[listofformat=parens, subrefformat=subparens]{subfig}
\usepackage{enumitem}
\renewcommand\cap[3]{\caption[#2]{\label{#1}\textsc{#2}. \small\textit{#3}}}
\renewcommand\etal[1]{~\textit{et\,al.}~\cite{#1}}

\usepackage[pagebackref=true,breaklinks=true,letterpaper=true,colorlinks,bookmarks=false]{hyperref}

\wacvfinalcopy % *** Uncomment this line for the final submission

\def\wacvPaperID{0430} % *** Enter the wacv Paper ID here

\ifwacvfinal\fi
\begin{document}

%%%%%%%%% TITLE
\title{Towards Robust Deep Neural Networks with BANG}

\author{Andras Rozsa, Manuel G\"unther, and Terrance E. Boult \\
Vision and Security Technology (VAST) Lab \\
University of Colorado, Colorado Springs, USA\\
\small\texttt{\{arozsa,mgunther,tboult\}@vast.uccs.edu}
}

\maketitle
\ifwacvfinal\thispagestyle{empty}\fi

\thispagestyle{fancy}
\fancyfoot[C]{}

\chead{\footnotesize This is a pre-print of the conference paper accepted at the IEEE Winter Conference on Applications of Computer Vision (WACV) 2018.}

%%%%%%%%% ABSTRACT
\begin{abstract}
Machine learning models, including state-of-the-art deep neural networks, are vulnerable to small perturbations that cause unexpected classification errors.
This unexpected lack of robustness raises fundamental questions about their generalization properties and poses a serious concern for practical deployments.
As such perturbations can remain imperceptible -- the formed adversarial examples demonstrate an inherent inconsistency between vulnerable machine learning models and human perception -- some prior work casts this problem as a security issue.
Despite the significance of the discovered instabilities and ensuing research, their cause is not well understood and no effective method has been developed to address the problem.
In this paper, we present a novel theory to explain why this unpleasant phenomenon exists in deep neural networks.
Based on that theory, we introduce a simple, efficient, and effective training approach, Batch Adjusted Network Gradients (BANG), which significantly improves the robustness of machine learning models.
While the BANG technique does not rely on any form of data augmentation or the utilization of adversarial images for training, the resultant classifiers are more resistant to adversarial perturbations while maintaining or even enhancing the overall classification performance.
\end{abstract}

%%%%%%%%% BODY TEXT
\section{Introduction}

Machine learning is broadly used in various real-world vision applications and recent advances in deep learning have made deep neural networks the most powerful learning models that can be successfully applied to different vision problems \cite{vinyals2015show,szegedy2015going,he2015deep,yang2015convolutional,ouyang2015deepid,lai2015simultaneous,long2015fully,lin2015deep,zhang2016efficient}. The recent performance gain is mainly the result of improvements in two fields, namely, building more powerful learning models \cite{szegedy2015going, he2015deep} and designing better strategies to avoid overfitting \cite{srivastava2014dropout}. %, which are then leveraged by larger datasets and massive GPU-enhanced computing.
These advancements are then leveraged by the use of larger datasets and massive GPU-enhanced computing.

\begin{figure}[!t]
\begin{center}
\centering\subfloat[][\label{fig:adv:a} MNIST Samples and Their Distortions Yielding Misclassifications]{\includegraphics[width=.98\columnwidth]{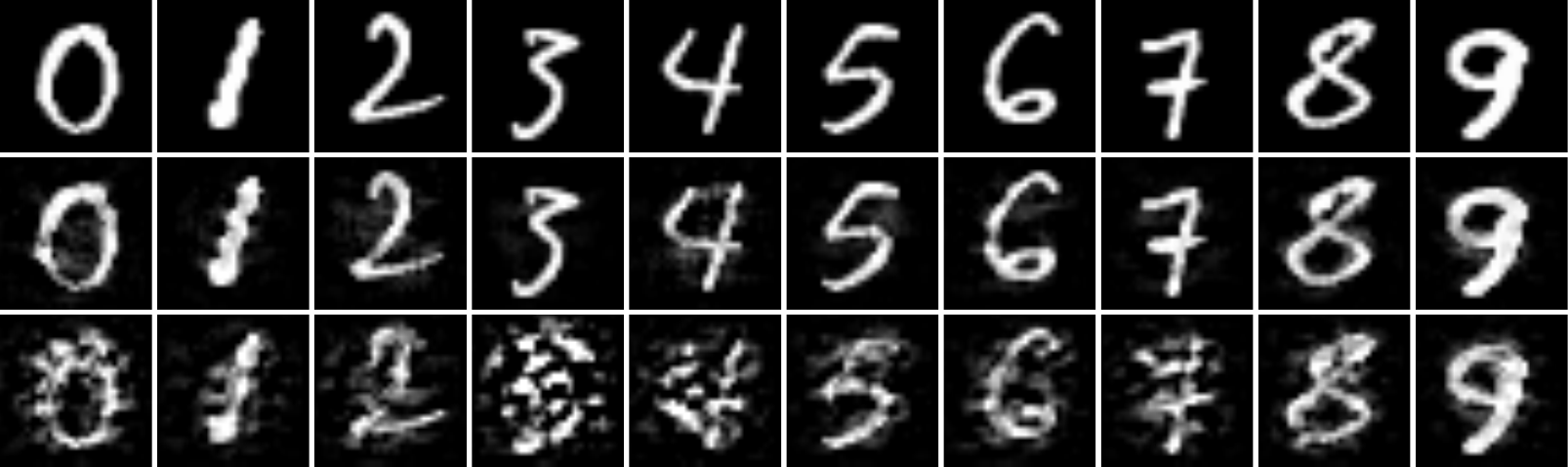}}
\vspace*{-0.06in}
\\
\centering\subfloat[][\label{fig:adv:b} CIFAR-10 Samples and Their Distortions Yielding Misclassifications]{\includegraphics[width=.98\columnwidth]{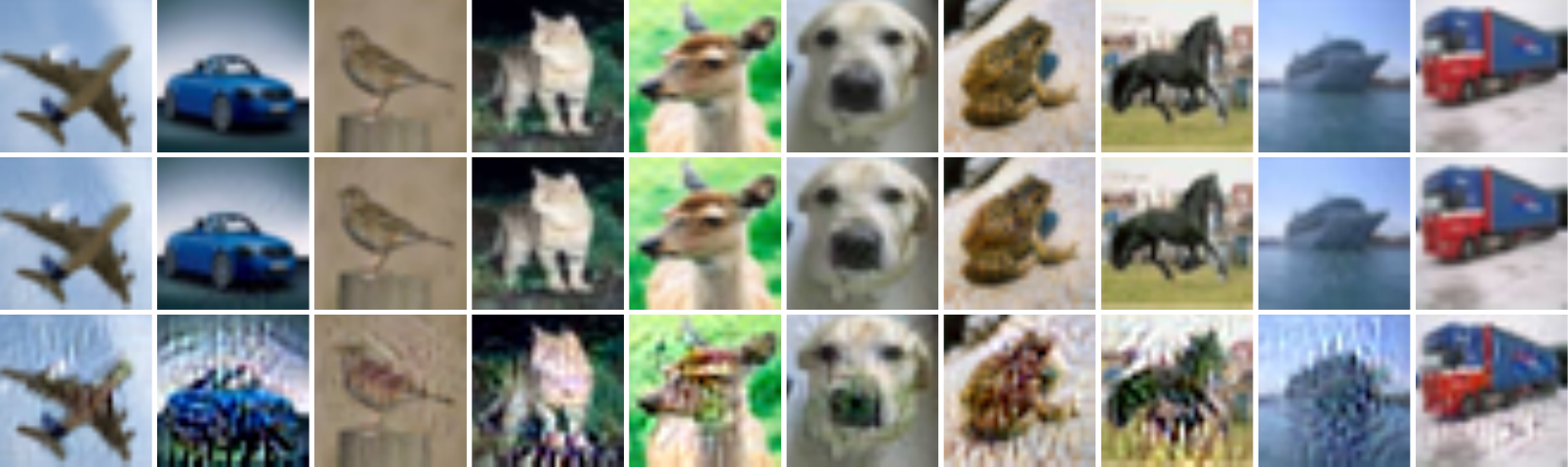}}
\vspace*{-0.06in}

\cap{fig:teaser}{Improving Robustness via BANG}{This figure demonstrates the enhanced robustness against perturbations generated via the non-gradient-based hot/cold adversarial generation method on MNIST digits and CIFAR-10 samples displayed in top rows of \subref*{fig:adv:a} and \subref*{fig:adv:b}. Underneath the raw test images, we show their distorted versions formed by the smallest perturbations that change the correctly classified class labels of the test samples. The second rows of \subref*{fig:adv:a} and \subref*{fig:adv:b} present perturbations that we obtained on regularly trained learning models, while the last rows show examples that we generated on networks trained via our Batch Adjusted Network Gradients (BANG) approach. As indicated by most of the perturbations being highly perceptible, the learning models trained with BANG have become more robust to adversarial perturbations.}
\vspace*{-0.18in}
\end{center}
\end{figure}

Although deep neural networks (DNNs) achieve state-of-the-art performance in a wide range of tasks, the generalization properties of these learning models were questioned by Szegedy\etal{szegedy2013intriguing} when the existence of adversarial examples was revealed.
DNNs are capable of learning high-level feature embeddings that enable them to be successfully adapted to different problems. They were generally considered to generalize well and, hence, expected to be robust to moderate distortions to their inputs.
Surprisingly, adversarial examples formed by applying imperceptible perturbations to otherwise correctly recognized inputs can lead machine learning models -- including state-of-the-art DNNs -- to misclassify those samples, often with high confidence.
This highly unexpected and intriguing property of machine learning models highlights a fundamental problem that researchers have been trying to solve.

To explain why adversarial examples exist, several controversial explanations were proposed.
As hypothesized in \cite{goodfellow2014explaining,fawzi2015fundamental}, adversarial instability exists due to DNNs acting as high-dimensional linear classifiers that allow even imperceptibly small, well-aligned perturbations applied to inputs to spread among higher dimensions and radically change the outputs.
This belief was challenged in \cite{luo2015foveation}, where -- by analyzing and experimenting with DNNs trained to recognize objects in more unconstrained conditions -- it was demonstrated that those classifiers are only locally linear to changes on the recognized object, otherwise DNNs act non-linearly.
After performing various experiments, Gu\etal{gu2015towards} concluded that adversarial instability is rather related to ``intrinsic deficiencies in the training procedure and objective function than to model topology.''

The problem addressed in this paper is not only about preventing attacks via adversarial examples, the focus is on the overall robustness and generalizability of DNNs.
This fundamental problem of deep learning has recently received increasing attention by researchers \cite{fawzi2016robustness,hein2017formal,peck2017lower}.
Considering state-of-the-art learning models applied to computer vision tasks, the classification of many incorrectly or uncertainly recognized inputs can be corrected and improved by small perturbations \cite{zheng2016improving,rozsa2016facial}, so this is a naturally occurring problem for learning-based vision systems.

In this paper, we introduce our theory on the instability of machine learning models and the existence of  adversarial examples: evolutionary stalling.
During training, network weights are adjusted using the gradient of loss, evolving to eventually classify examples correctly.
Ideally, we prefer broad flat regions around samples to achieve good generalization~\cite{keskar2016large} and adversarial robustness~\cite{fawzi2015fundamental}.
However, after a training sample is correctly classified, its contribution to the loss and, thus, on forming the weight updates is reduced.
As the evolution of the local decision surface stalls, the correctly classified samples cannot further flatten and extend their surroundings to improve generalization.
Therefore, as the contributions of those correctly classified training samples to boundary adjustments are highly decreased compared to other batch elements, samples can end up being stuck close to decision boundaries and, hence, susceptible to small perturbations flipping their classifications.

To mitigate evolutionary stalling, we propose our Batch Adjusted Network Gradients (BANG) training algorithm.
We experimentally evaluate robustness using a combination of gradient- and non-gradient-based adversarial perturbations, and random distortions.
The paper explores the impact of BANG parameters and architectural variations, such as Dropout~\cite{srivastava2014dropout}, on instability and adversarial robustness.
In conclusion, we validate our theory by experimentally demonstrating that BANG significantly improves the robustness of deep neural networks optimized on two small datasets while the trained learning models maintain or even improve their overall classification performance.

\section{Related Work}

Deep neural networks (DNNs) achieve high performance on various tasks as they are able to learn non-local generalization priors from training data. Counter-intuitively, Szegedy\etal{szegedy2013intriguing} showed that machine learning models can misclassify samples that are formed by slightly perturbing correctly recognized inputs. These so-called adversarial examples are indistinguishable from their originating counterparts to human observers, and their unexpected existence itself presents a problem. The authors introduced the first technique that is capable of reliably finding adversarial perturbations and claimed that some adversarial examples generalize across different learning models.

A computationally cheaper adversarial example generation algorithm, the Fast Gradient Sign (FGS) method, was presented by Goodfellow\etal{goodfellow2014explaining}.
While this approach also uses the inner state of DNNs, it is more efficient as FGS requires the gradient of loss to be calculated only once.
The authors demonstrated that by using adversarial examples generated with FGS implicitly in an enhanced objective function, both accuracy and robustness of the trained classifiers can be improved.
In their paper focusing on adversarial machine learning, Kurakin\etal{kurakin2017adversarial} proposed new algorithms extending the FGS method to target a specific class and to calculate and apply gradients iteratively instead of a single gradient calculation via FGS.
The authors compared the effect of  different types of adversarial examples used for implicit adversarial training and  found that the results vary based upon the type of the applied adversarial examples.

Rozsa\etal{rozsa2016adversarial} introduced the non-gradient-based hot/cold approach, which is capable of efficiently producing multiple adversarial examples for each input. They demonstrated that using samples explicitly with higher magnitudes of adversarial perturbations than the sufficient minimal can outperform regular adversarial training. The authors also presented a new metric -- the Perceptual Adversarial Similarity Score (PASS) -- to better measure the distinguishability of original and adversarial image pairs in terms of human perception. As the commonly used L$_2$ or L$_\infty$ norms are very sensitive to small geometric distortions that can remain unnoticeable to us, PASS is more applicable to quantify similarity and the quality of adversarial examples.

Although adversarial training, both implicit and explicit, was demonstrated to decrease the instability of learning models, forming those examples is still computationally expensive, which limits the application of such techniques. Furthermore, considering the various adversarial generation techniques, utilizing certain types of those samples might not lead to improved robustness to adversarial examples of other techniques. Alternatively, Zheng\etal{zheng2016improving} proposed their stability training as a lightweight and still effective method to stabilize DNNs against naturally occurring distortions in the visual input. The introduced training procedure uses an additional stability objective that makes DNNs learn weights that minimize the prediction difference of original and perturbed images. In order to obtain general robustness and not rely on any class of perturbations, the authors applied Gaussian noise to distort the training images.

Gu\etal{gu2015towards} conducted experiments with different network topologies, pre-processing, and training procedures to improve the robustness of DNNs. The authors proposed the Deep Contractive Network (DCN), which imposes a layer-wise contractive penalty in a feed-forward DNN. The formulated penalty aims to minimize output variances with respect to perturbations in inputs, and enable the network to explicitly learn flat, invariant regions around the training data. Based on positive initial results, they concluded that adversarial instability is rather the result of the intrinsic deficiencies in the training procedure and objective function than of model topologies.

Luo\etal{luo2015foveation} proposed a foveation-based technique that selects and uses only a sub-region of the image during classification. As the authors demonstrated, the negative effect of foveated perturbations to the classification scores can be significantly reduced compared to entire perturbations. Graese\etal{graese2016assessing} showed that transformations of the normal image acquisition process can also negate the effect of the carefully crafted adversarial perturbations. While these pre-processing techniques can alleviate the problem posed by adversarial images, they do not solve the inherent instability of DNNs. In other words, these methods treat the symptoms and not the disease.

In summary, a wide variety of more or less efficient approaches were proposed in the literature that all aim at improving the robustness and generalization properties of DNNs, but none of those proved to be effective enough.

\section{Approach}

In this section, we first briefly describe our intuition about why the unexpected adversarial instability exists in machine learning models. Afterwards, we present our simple and straightforward modification in the training procedure that aims to optimize weights in a way that the resulting DNNs become more robust to distortions of their inputs.

\subsection{Intuition}

During training, some inputs in the batch are correctly and others are incorrectly classified.
In general, the calculated loss and, thus, the gradient of loss for the misclassified ones are larger than for the correctly classified inputs of the same batch.
Therefore, in each training iteration most of the weight updates go into learning those inputs that are badly predicted.
On the other hand, the correctly classified samples do not have a significant impact on advancing decision boundaries and can remain in the positions close to what they obtained when becoming correctly classified.
Due to this evolutionary stalling, samples with low gradients cannot form a flatter, more invariant region around themselves.
Consequently, samples of those regions remain more susceptible to adversarial perturbations -- even a small perturbation can push them back into an incorrect class.
By increasing the contribution of the correctly classified examples in the batch on the weight updates, and forcing them to continue improving decision boundaries, it is reasonable to think that we can flatten the decision space around those training samples and train more robust DNNs.

\subsection{Implementation}

The core concept of our Batch Adjusted Network Gradients (BANG) approach is a variation of batch normalization \cite{ioffe2015batch}. However, rather than trying to balance the inputs of the layers, we seek to ensure that the contributions on the weight updates are more balanced among batch elements by scaling their gradients.

Let us dive into the details and introduce our notations we use to formulate BANG. In short, we scale the gradients of batch elements that will be used to compute the weight updates in each training iteration.
Let us consider a network $f_w$ with weights $w$ in a layered structure having layers $y^{(l)}$ where $l\in\left[1,L\right]$, with their respective weights $w^{(l)}$:
\begin{equation}
  %  f_w(x_i) = y^{(L)}\left(y^{(L-1)}\left(\dots\left(y^{(1)}(x_i)\right)\dots\right)\right).
    f_w(x_i) = y^{(L)}\left(y^{(L-1)}\left({\dots{\left(y^{(1)}(x_i)\right)\dots}}\right)\right).
\end{equation}
For a given input $x_i$, the partial derivatives of the loss $E(f_w, x_i)$ with respect to the output of layer $y^{(l)}$ are:
\begin{equation}
  \kappa_i^{(l)} = \kappa^{(l)}(x_i) = \frac{\partial E_i}{\partial y_i^{(l)}}.
\end{equation}
%For simplicity, we leave out the structure of the weights $w^{(l)}$ in each layer which are commonly in a two-dimensional structure, and of the layer output which are either one-dimensional (fully connected layers) or three-dimensional (convolutional layers).
For simplicity, we leave out the structure of the weights $w^{(l)}$ in layers and the structure of the layer outputs which can be either one-dimensional for fully connected layers or three-dimensional for convolutional layers.

With BANG, our goal is to balance gradients in the batch by scaling up those that have lower magnitudes. In order to do so, we determine the highest gradient for the batch having $N$ inputs $x_i, i\in \{1,\ldots,N\}$ at given layer $y^{(l)}$ in terms of $L_2$ norm.
We use that as the basis for balancing the magnitudes of gradients in the batch.
Weight updates are calculated after scaling each derivative $\kappa_i$ in the batch with the element-wise learning rate:
\begin{equation}
  \eta_i^{(l)} = \left(\frac{\max\limits_{i' \in [1,N]} \|\kappa_{i'}^{(l)}\|} {\|\kappa_i^{(l)} \|} \right) ^ {\rho_i^{(l)}}\,
\end{equation}
where:
\begin{equation}
\label{eq:exp}
  \rho_i^{(l)} = \epsilon^{(l)} \left(1-\frac{\|\kappa_i^{(l)} \|}{\max\limits_{i' \in [1,N]}\|\kappa_{i'}^{(l)} \|} \right).
\end{equation}
As a key parameter for our approach, $\epsilon^{(l)}$ specifies the degree of gradient balancing among batch elements.
While the exponent $\rho_i^{(l)}$ might appear a little complex and ambiguous, its sole purpose is to scale up gradients with small magnitudes more than others having larger $L_2$ norms.

Assuming that the regular backward pass combines the gradients of the batch elements by calculating:
\begin{equation}
  \nabla f_{w^{(l)}} = \frac{1}{N} \sum\limits_{i=1}^N \frac{\partial E_i}{\partial w^{(l)}} = \frac{1}{N} \sum\limits_{i=1}^N \kappa_i^{(l)} \frac{\partial y^{(l)}}{\partial w^{(l)}}
\end{equation}
which is normally scaled with the learning rate and then used to update weights (after combining with the previous weight update scaled with momentum), BANG produces:
\begin{equation}
\label{eq:update}
  \nabla f_{w^{(l)}} = \beta^{(l)} \frac{1}{N} \sum\limits_{i=1}^N \eta_i^{(l)} \frac{\partial E_i}{\partial w^{(l)}},
\end{equation}
where $\beta^{(l)}$ is the second (set of) parameter(s) of our approach used for scaling. In general, $\beta^{(l)}$ acts as a local learning rate that can play a more important role in future work.
Throughout our experiments, we keep BANG parameters fixed for all layers: $\epsilon^{(l)} = \epsilon$ and $\beta^{(l)} = \beta$ (which will actually just modify the original learning rate $\eta$).

Note that although our approach changes the actual calculation of weight updates for the layers, there is no impact on the backpropagation of the original gradient down the network. Finally, we implemented BANG by applying small modifications to the regular training procedure with negligible computational overhead.

\newcommand\cen[1]{\multicolumn{1}{c}{#1}}

\section{Experiments}

To evaluate our approach, we conducted experiments on the slightly modified versions of LeNet \cite{lecun1995learning} and ``CIFAR-10 quick'' models distributed with Caffe~\cite{jia2014caffe}.
Namely, after running preliminary experiments with BANG, we added a Dropout layer \cite{srivastava2014dropout} to both model architectures that serves multiple purposes. We observed that BANG tends to cause overfitting on the trained LeNet networks, and the resultant models made very confident classifications -- even when they misclassified the test images.
While the additional Dropout layer alleviates both problems, the adjusted network architectures also result in improved classification performances with both regular and BANG training.

After obtaining learning models with regular and BANG training, we assess and compare the robustness of those classifiers in two ways. It is important to note that we do not select the best training models based on their performance on the validation set for these evaluations, but we simply use the models obtained at the last training iteration. As our primary goal is to measure the evolving robustness, we believe that this decision leads to a fairer comparison, however, the classification performance of the selected models are not optimal. Finally, we would like to mention that we conducted experiments to discover the effectiveness of BANG used for fine-tuning regularly trained models, and found that the robustness of the resultant networks are not even comparable to those that we trained from scratch.

\begin{table*}
\setlength{\tabcolsep}{7pt}
\scriptsize
\vspace*{0.05in}
\cap{tab:lenets}{LeNet Training}{This table highlights the difference between LeNet models obtained by using regular (R0-R1) and BANG training (B0-B5). Accuracy on the MNIST test set, the achieved success rates of FGS and HC1 adversarial example generation methods with PASS scores and L$_\infty$ norms of the produced examples on the MNIST test set are listed.}
\vspace*{-0.15in}
\begin{center}
  \begin{tabular}{ccccrcrccr}
    \toprule
    ID	 &$\beta$ &$\epsilon$ & Accuracy & \cen{FGS-Rate} & FGS-PASS & \cen{FGS-L$_\infty$} & HC1-Rate & HC1-PASS & \cen{HC1-L$_\infty$}\\
    \midrule
    \midrule
    R0	& -	& -	& $99.16$\% & $90.33$\%\, & $0.4072\pm0.1081$ & $40.51\pm15.72$ & $99.53$\% & $0.7535\pm0.1143$ & $122.60\pm49.46$\\
    \midrule
    R1	& -	& -	& $99.15$\% & $91.41$\%\, & $0.4072\pm0.1065$ & $40.70\pm15.88$ & $99.77$\% & $0.7517\pm0.1160$ & $122.16\pm49.07$\\
    \midrule
    B0	& $1.00$	& $0.785$ & $99.16$\% & $3.51$\%\, & $0.6806\pm0.1457$ & $8.34\pm04.32$ & $95.13$\% & $0.5359\pm0.2023$ & $187.86\pm63.66$\\
    \midrule
    B1	& $1.00$	& $0.815$ & $99.22$\% & $1.68$\%\, & $0.7638\pm0.1367$ & $5.52\pm02.86$ & $94.19$\% & $0.4880\pm0.2110$ & $201.84\pm62.51$\\
    \midrule
    B2	& $1.20$	& $0.810$ & $99.31$\% & $2.13$\%\, & $0.7579\pm0.1452$ & $5.57\pm03.05$ & $94.56$\% & $0.5129\pm0.2015$ & $186.10\pm63.77$\\
    \midrule
    B3	& $1.35$	& $0.780$ & $99.25$\% & $3.86$\%\, & $0.6763\pm0.1471$ & $8.28\pm04.26$ & $94.73$\% & $0.5709\pm0.2127$ & $178.11\pm65.76$\\
    \midrule
    B4	& $1.50$	& $0.840$ & $99.11$\% & $1.52$\%\, & $0.8220\pm0.1310$ & $4.19\pm03.01$ & $97.68$\% & $0.4669\pm0.1881$ & $203.50\pm58.97$\\
    \midrule
    B5	& $1.60$	& $0.780$ & $99.32$\% & $4.45$\%\, & $0.6771\pm0.1487$ & $8.20\pm04.40$ & $98.95$\% & $0.6376\pm0.1829$ & $146.96\pm61.97$\\
    \bottomrule
  \end{tabular}
\end{center}
\end{table*}

\begin{figure*}
\begin{center}
\vspace*{-0.2in}
\centering\subfloat[][\label{fig:bang_lenets:a}\centering Accuracy]{\includegraphics[width=0.32\linewidth]{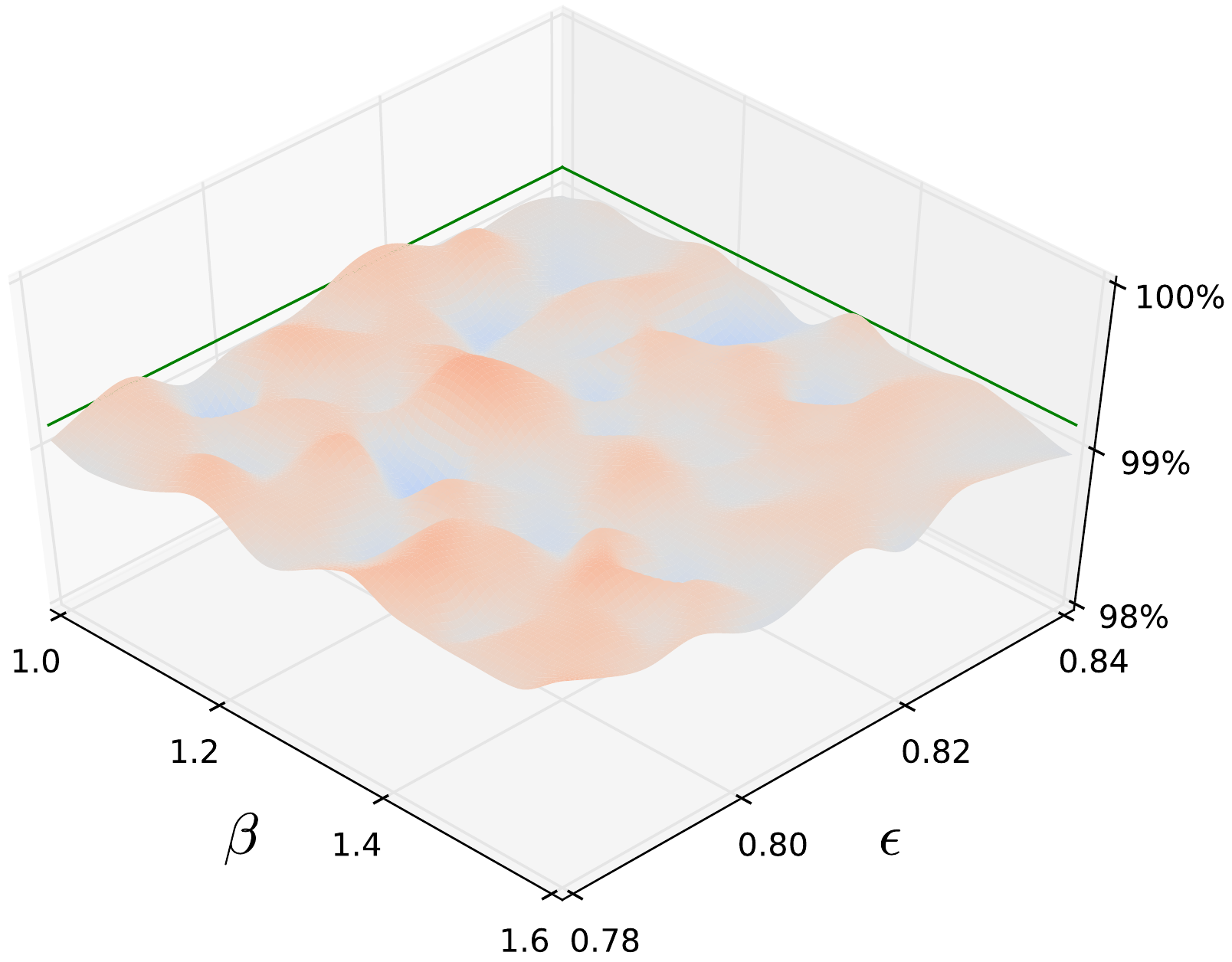}}
\hfill
\centering\subfloat[][\label{fig:bang_lenets:b}\centering FGS Success Rate]{\includegraphics[width=0.32\linewidth]{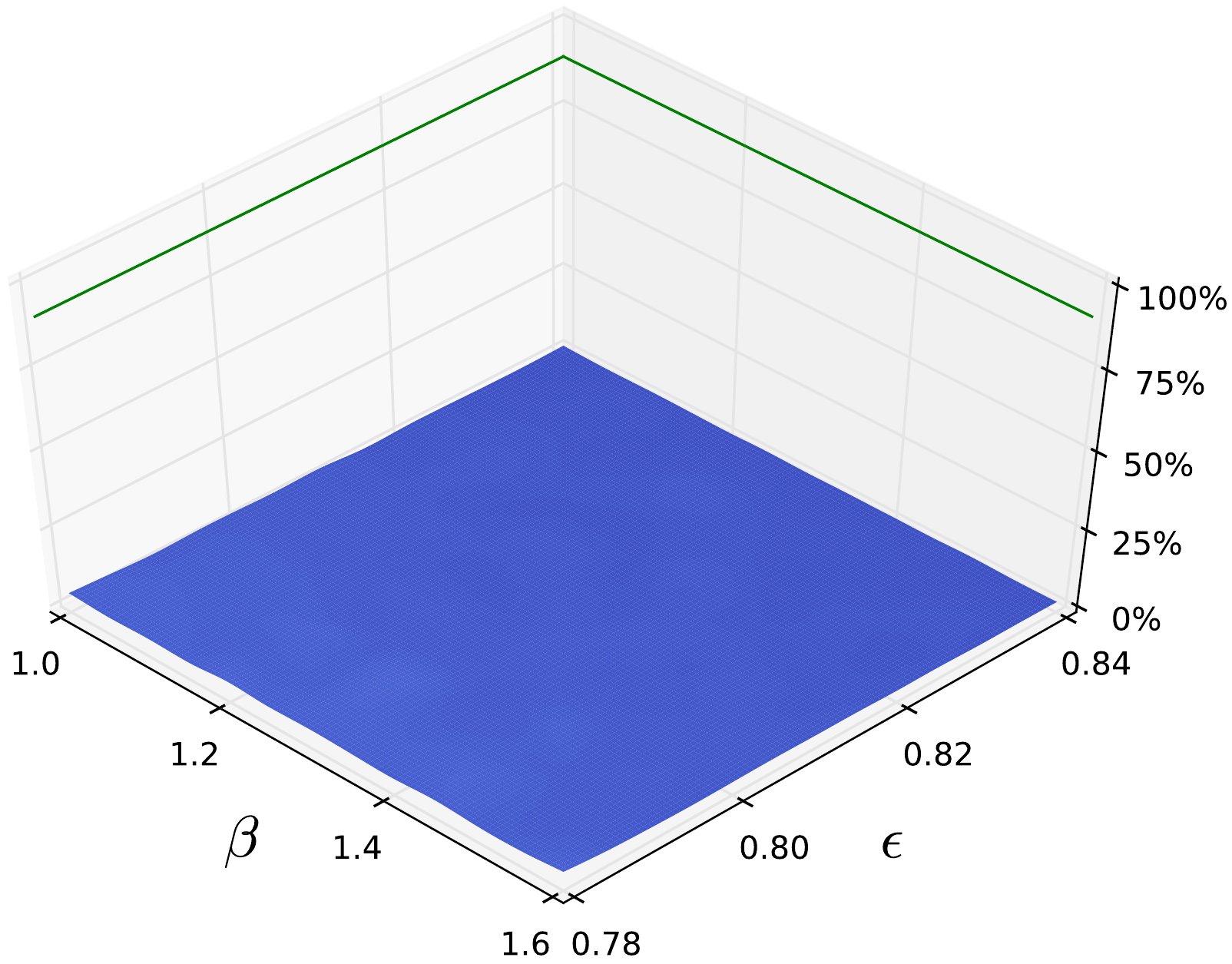}}
\hfill
\centering\subfloat[][\label{fig:bang_lenets:c}\centering HC1 PASS]{\includegraphics[width=0.32\linewidth]{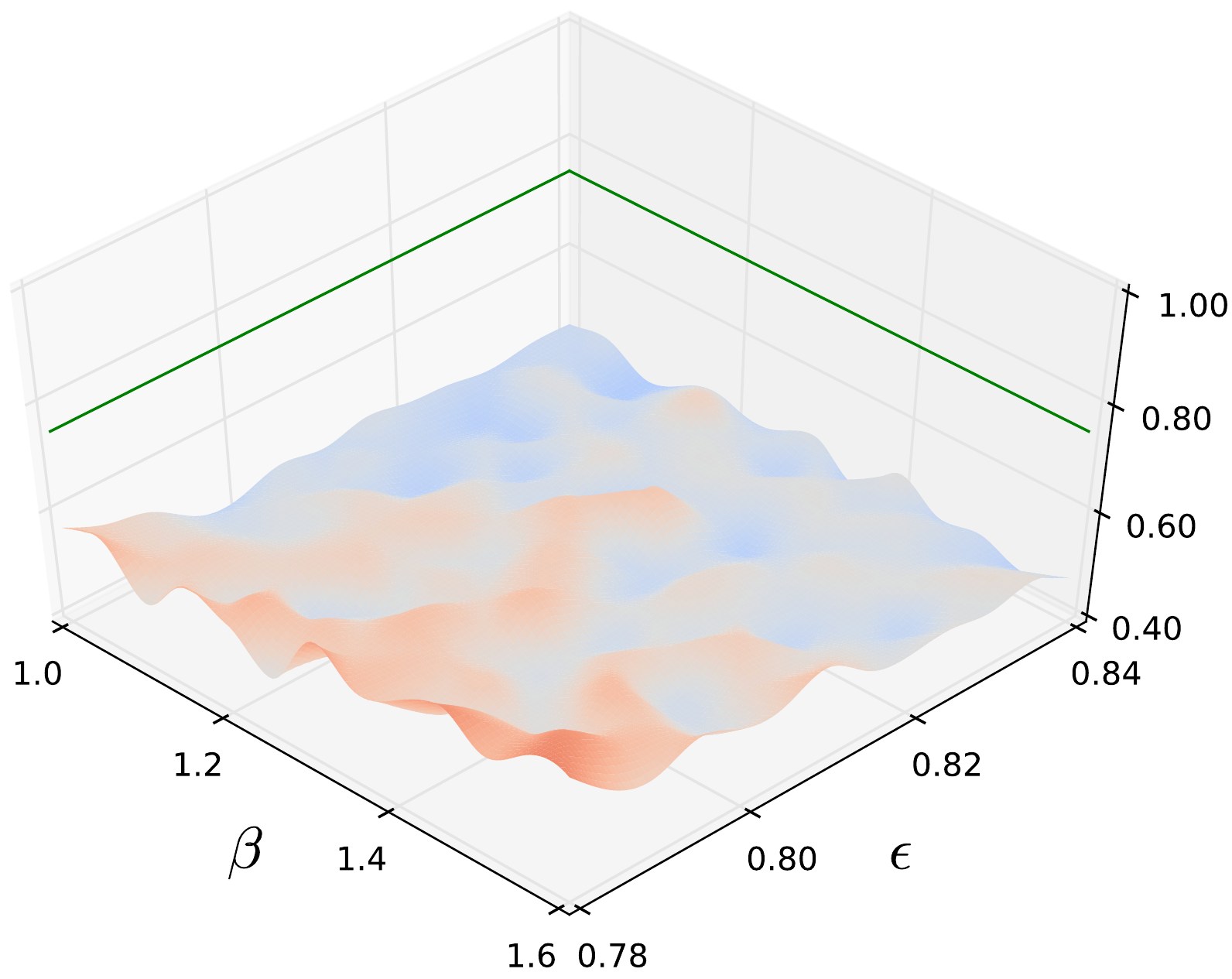}}
\vspace*{-0.05in}
\cap{fig:bang_lenets}{LeNet Models Trained with BANG}{These plots summarize our results on LeNet models trained with BANG using combinations of $\beta$ and $\epsilon$. We tested a grid of those two parameters where $\beta\in\left[1.0,1.6\right]$ with step size $0.05$, and $\epsilon\in\left[0.78,0.84\right]$ with step size $0.005$. We trained a single model with each combination and show \subref*{fig:bang_lenets:a} the obtained accuracy on the MNIST test set, \subref*{fig:bang_lenets:b} the achieved success rates by using FGS and \subref*{fig:bang_lenets:c} the mean PASS score of HC1 adversarial examples on the MNIST test images. Each solid green line represents the level of regularly trained learning models. For better visual representation we applied interpolation.}
\end{center}
\vspace*{-0.19in}
\end{figure*}

First, we evaluate the adversarial vulnerability against two adversarial example generation methods: the gradient-based Fast Gradient Sign (FGS) method \cite{goodfellow2014explaining} and the non-gradient-based hot/cold approach \cite{rozsa2016adversarial}. Although the latter is capable of forming multiple adversarial perturbations for each input, we only target the most similar class with the hot/cold approach, referred to as HC1.

We aim to form adversarial perturbations for every correctly classified image from the MNIST~\cite{lecun1998mnist} or CIFAR-10~\cite{krizhevsky2009learning} test set, respectively.
We consider an adversarial example generation attempt successful, if the direction specified by either FGS or HC1 leads to a misclassification, where the only constraint is that the discrete pixel values are in $\left[0,255\right]$ range. Of course, this limitation means that the formed perturbations may or may not be adversarial in nature as they can be highly perceptible to human observers.
We compare the adversarial robustness of classifiers by collecting measures to quantify the quality of the produced adversarial examples. For this purpose, we calculate the Perceptual Adversarial Similarity Score (PASS)~\cite{rozsa2016adversarial} of original and adversarial image pairs, and we also determine the $L_\infty$ norms of adversarial perturbations. Although the $L_\infty$ norm is not a good metric to quantify adversarial quality in terms of human perception, it can demonstrate how far the actual perturbed image is from the original sample.

Second, we quantify how the robustness of the learning models evolve during training by applying a more general approach. For a given pair of classifiers where one was regularly trained while the other was obtained by BANG training, we add a certain level of random noise to 100 test images from each class that are correctly classified by both networks at all tested stages and compute the proportion of perturbed images that are classified differently than the originating one. While the previously described test assessing the adversarial vulnerability explores only two directions -- specified by the FGS method and the HC1 approach -- applying 1000 random distortions to each inspected image for every noise level gives us a more general evaluation.

Although experimenting with random noise is more universal as it does not rely on any specific adversarial generation technique, small random perturbations that cause misclassifications are hard to find \cite{rozsa2016adversarial} and, hence, the collected results are qualitatively not as good as explicitly forming adversarial perturbations.
Furthermore, in order to evaluate the stability of the trained classifiers, we distorted the images with Gaussian noise far beyond the noise level that can be considered imperceptible or adversarial.

\subsection{LeNet on MNIST}

We commenced our experiments by evaluating BANG on the LeNet model optimized on the MNIST dataset. MNIST contains 70k images overall: 50k used for training, 10k for validation, and the remaining 10k for testing.
The tested network originally has four layers (two convolutional and two fully connected) -- extended with one additional Dropout layer -- that we optimize without changing the hyperparameters distributed with Caffe. The learning model is trained with a batch size of 64 for 10k iterations using the inverse decay learning rate policy with an initial learning rate of 0.01.

\begin{figure*}
\begin{center}
\vspace*{-0.06in}
\centering\subfloat[][\label{fig:lenets_noise:a}\centering Regular Training]{\includegraphics[width=0.32\linewidth]{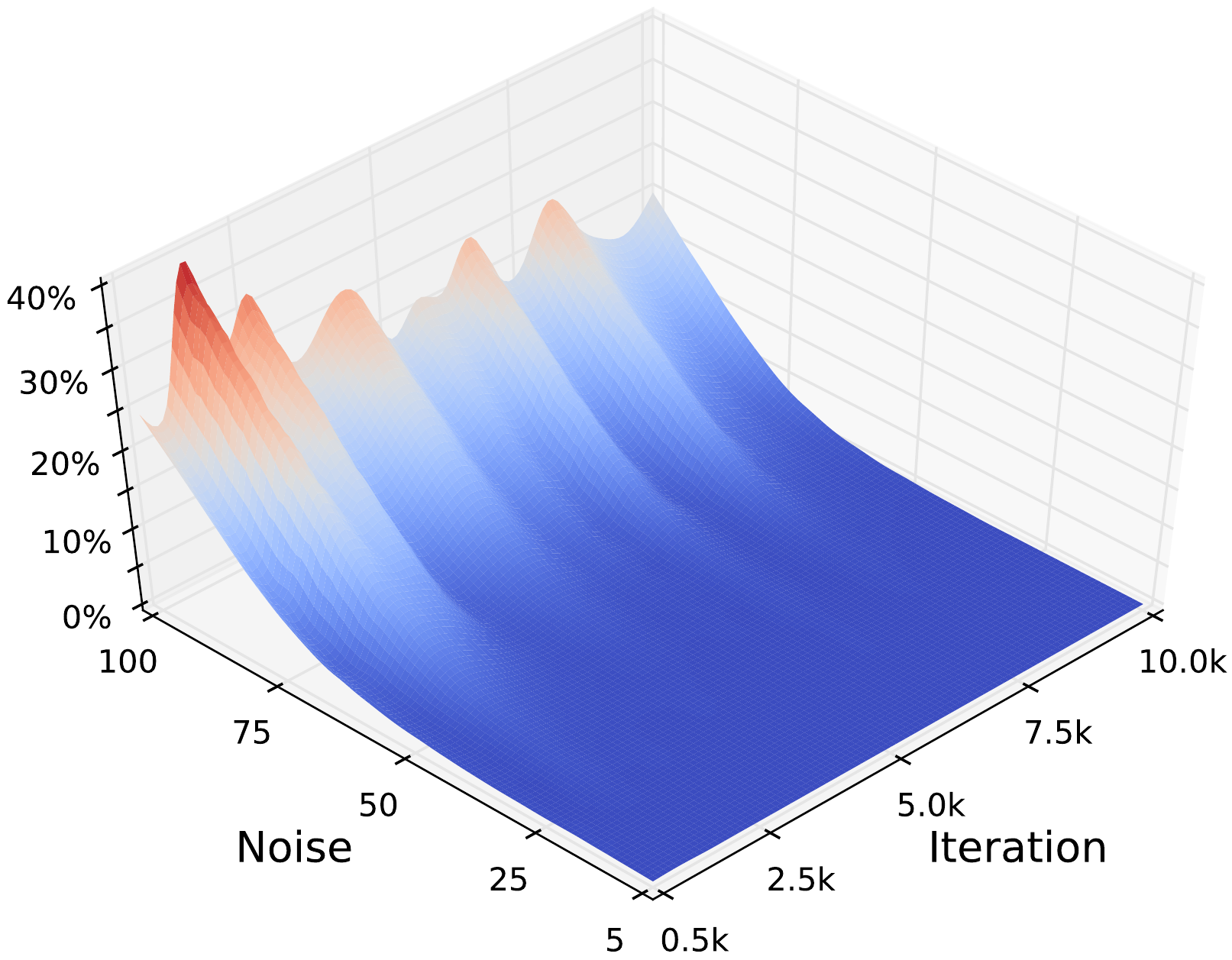}}
\hfill
\centering\subfloat[][\label{fig:lenets_noise:b}\centering BANG Training]{\includegraphics[width=0.32\linewidth]{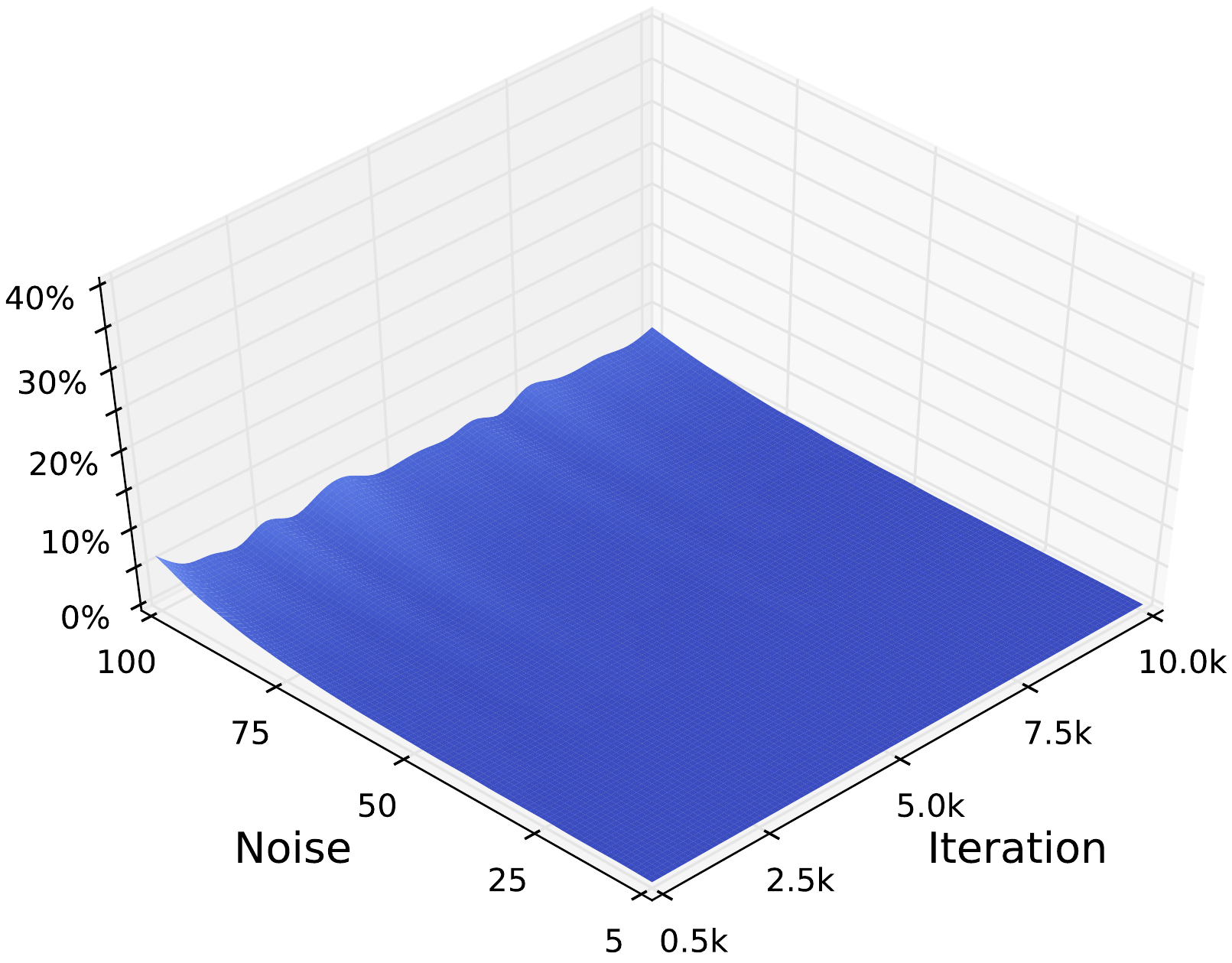}}
\hfill
\centering\subfloat[][\label{fig:lenets_noise:c}\centering Absolute Improvement]{\includegraphics[width=0.32\linewidth]{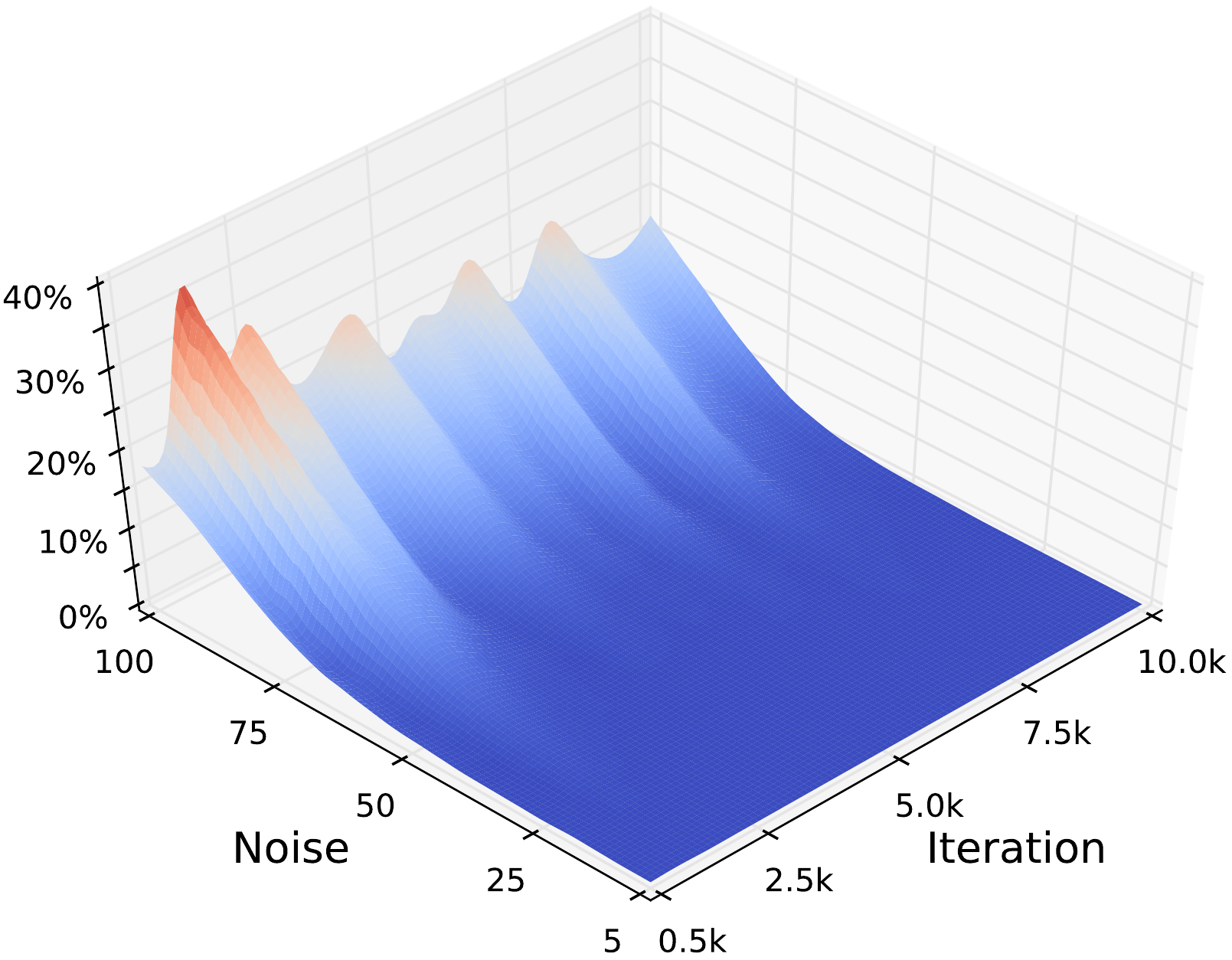}}
\vspace*{-0.05in}
\cap{fig:lenets_noise}{LeNet: Robustness to Random Distortions}{These plots show the evolving robustness of LeNet models: \subref*{fig:lenets_noise:a} obtained with regular training (R0 from Table~\ref{tab:lenets}), \subref*{fig:lenets_noise:b} trained with BANG (B1 from Table~\ref{tab:lenets}), and \subref*{fig:lenets_noise:c} displays the improvement. After identifying 100 test images per class that are correctly classified by both networks at every 500 iterations, we perturb each 1000 times by adding the level of Gaussian noise specified by the standard deviation, and test the networks at several stages of training. The plots show the percentage of distortions yielding misclassifications. For better visual representation we applied interpolation.}
\end{center}
\vspace*{-0.05in}
\end{figure*}

Since our training procedure has two parameters, $\beta$ defined in Equation~\eqref{eq:update} and $\epsilon$ introduced in Equation~\eqref{eq:exp}, we trained LeNet models with parameter combinations from a grid, and evaluated the accuracy and adversarial vulnerability of the trained classifiers. The results of the conducted experiments are visualized in Figure~\ref{fig:bang_lenets}, we also show accuracies and metrics indicating adversarial robustness in Table~\ref{tab:lenets} for some models obtained with regular training (R0-R1) and optimized with BANG training (B0-B5).

As we can see in Table~\ref{tab:lenets}, FGS success rates achieved by regular training can be dramatically decreased by BANG: the rate drops from above 90\% to below 2\%. Almost every single failed adversarial example generation attempt is due to blank gradients -- the gradient of loss with respect to the original image and its ground-truth label contains only zeros -- which means that methods utilizing that gradient of loss cannot succeed.
As we increase $\epsilon$, or in other words, as we balance the contributions of batch elements more by scaling up gradients with lower magnitudes, the resultant classifiers become more resistant to gradient-based adversarial generation methods. Although the success rates obtained by the HC1 method remain relatively high, the qualities of HC1 examples degrade significantly on LeNet models trained with BANG compared to the regular training as displayed in Figure~\subref{fig:bang_lenets:c}. This degradation is highlighted by both decreasing PASS scores and by the significantly increased $L_\infty$ norms of perturbations listed in Table~\ref{tab:lenets}.

With respect to the achieved classification performances, we find that there can be a level of degradation depending on the selected values for $\beta$ and $\epsilon$. This phenomenon can be seen in Figure~\subref{fig:bang_lenets:a}; it is partially due to random initializations and can be the result of overfitting or our decision to evaluate all networks at 10k training iterations. Still, we can observe that BANG can yield improved classification performance over regular training paired with improved robustness as listed in Table~\ref{tab:lenets}.

Additionally, we conducted experiments to quantify and compare how the robustness to random perturbations evolves during training. For this general approach, we selected to test two classifiers from Table~\ref{tab:lenets}: R0 optimized with regular training and B1 trained with BANG. We can see in Figure~\subref{fig:lenets_noise:a} that the regularly trained model is initially highly susceptible to larger distortions, but as the training progresses it becomes more stable, and settles at approximately 20\% with respect to the strongest class of Gaussian noise that we formed by using standard deviation of 100 pixels. Contrarily, the classifier trained with BANG maintains significantly lower rates throughout the whole training as shown in Figure~\subref{fig:lenets_noise:b}, and after 10k iterations only 3\% of the strongest distortions can alter the original classification. The absolute improvements are displayed in Figure~\subref{fig:lenets_noise:c}.

\subsection{CIFAR-10}

\begin{table*}
\setlength{\tabcolsep}{7pt}
\scriptsize
\vspace*{0.05in}
\cap{tab:cifar10s}{CIFAR-10 Training}{This table shows the difference between classifiers obtained using regular (R0-R1) and BANG training (B0-B5). The accuracy on the CIFAR-10 test set, the achieved success rates of FGS and HC1 adversarial example generation methods with PASS scores and L$_\infty$ norms of the formed examples on the CIFAR-10 test images are listed.}
\label{cifar10-table}
\vspace*{-0.15in}
\begin{center}
  \begin{tabular}{ccccrcrccr}
    \toprule
    ID	 &$\beta$ &$\epsilon$ & Accuracy & \cen{FGS-Rate} & FGS-PASS & \cen{FGS-L$_\infty$} & HC1-Rate & HC1-PASS & \cen{HC1-L$_\infty$}\\
    \midrule
    \midrule
    R0	& -	& -	& $79.59$\% & $96.52$\%\, & $0.9553\pm0.0969$ & $4.08\pm06.40$ & $98.97$\% & $0.9669\pm0.1005$ & $18.15\pm29.80$\\
    \midrule
    R1	& -	& -	& $79.55$\% & $96.71$\%\, & $0.9513\pm0.1057$ & $4.43\pm07.05$ & $98.91$\% & $0.9557\pm0.1332$ & $22.16\pm39.77$\\
    \midrule
    B0	& $0.40$	& $0.855$ & $79.26$\% & $34.27$\%\, & $0.9511\pm0.1302$ & $4.11\pm10.31$ & $95.94$\%	& $0.8712\pm0.1649$ & $55.52\pm49.98$\\
    \midrule
    B1	& $0.45$	& $0.805$ & $80.43$\% & $45.94$\%\, & $0.9818\pm0.0548$ & $2.04\pm02.49$ & $96.20$\%	& $0.7966\pm0.2438$ & $77.34\pm71.20$\\
    \midrule
    B2	& $0.75$	& $0.800$ & $79.74$\% & $41.71$\%\, & $0.9828\pm0.0586$ & $1.94\pm03.03$ & $98.34$\%	& $0.8362\pm0.2195$ & $64.26\pm63.57$\\
    \midrule
    B3	& $0.75$	& $0.845$ & $79.41$\% & $35.00$\%\, & $0.9526\pm0.1266$ & $3.94\pm08.71$ & $96.54$\% & $0.8603\pm0.1981$ & $59.83\pm58.28$\\
    \midrule
    B4	& $0.95$	& $0.840$ & $79.30$\% & $34.88$\%\, & $0.9575\pm0.1236$ & $3.61\pm09.60$ & $96.87$\% & $0.8994\pm0.1487$ & $48.44\pm47.35$\\
    \midrule
    B5	& $1.00$	& $0.800$ & $79.22$\% & $41.34$\%\, & $0.9803\pm0.0722$ & $2.03\pm03.64$ & $98.17$\% & $0.8586\pm0.1948$ & $61.14\pm61.23$\\
    \bottomrule
  \end{tabular}
\end{center}
\end{table*}

\begin{figure*}
\begin{center}
\vspace*{-0.2in}
\centering\subfloat[][\label{fig:bang_cifar10s:a}\centering Accuracy]{\includegraphics[width=0.32\linewidth]{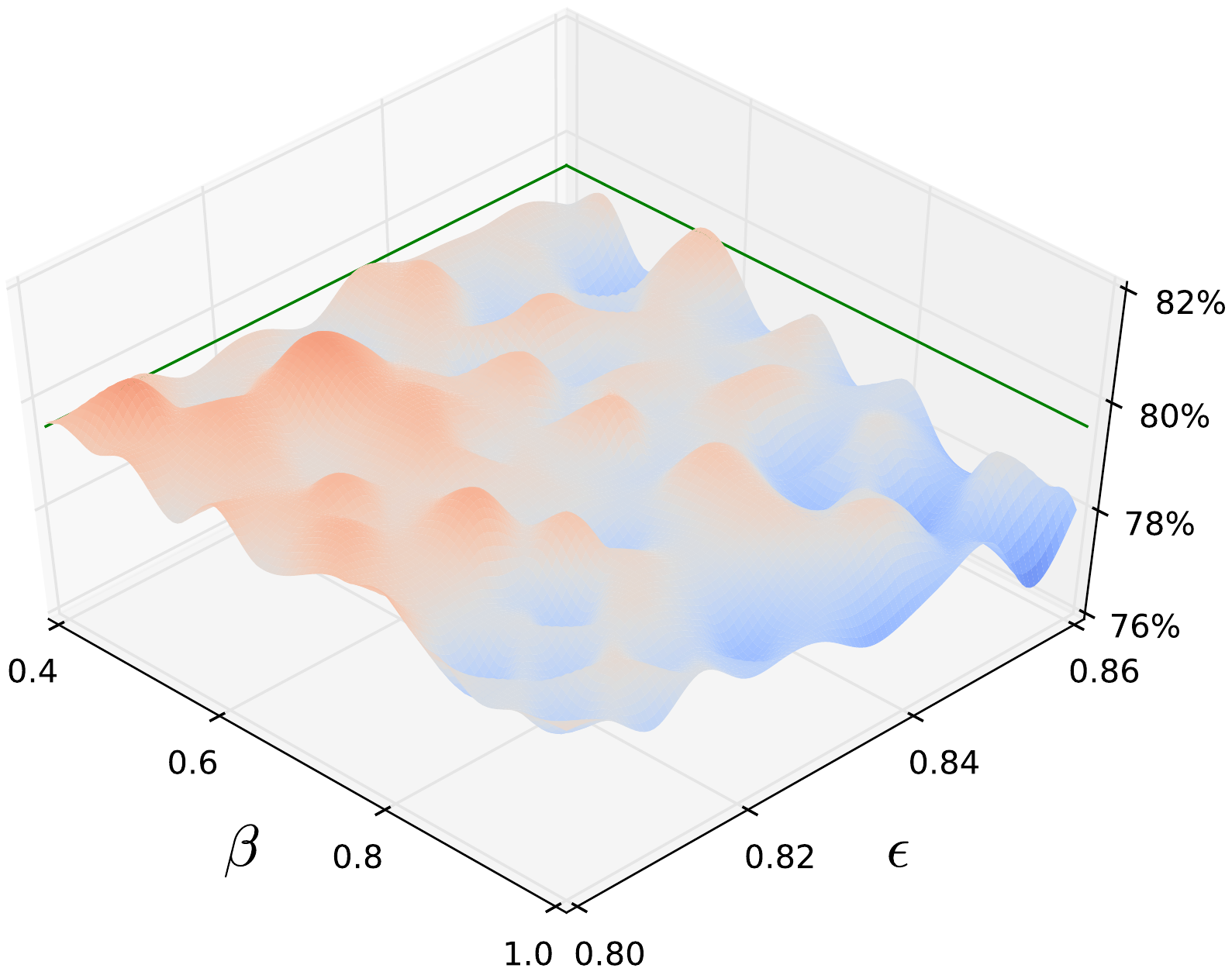}}
\hfill
\centering\subfloat[][\label{fig:bang_cifar10s:b}\centering FGS Success Rate]{\includegraphics[width=0.32\linewidth]{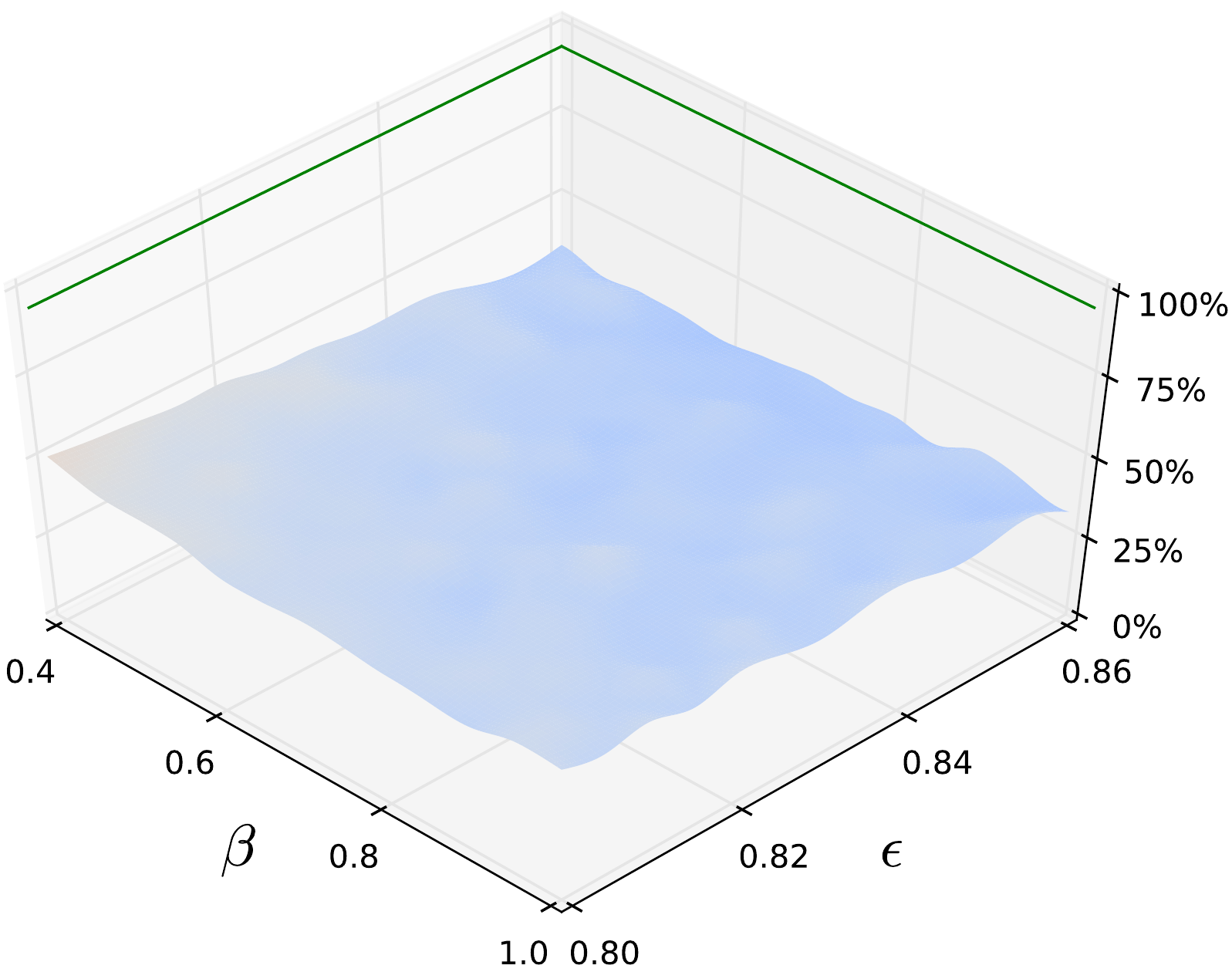}}
\hfill
\centering\subfloat[][\label{fig:bang_cifar10s:c}\centering HC1 PASS]{\includegraphics[width=0.32\linewidth]{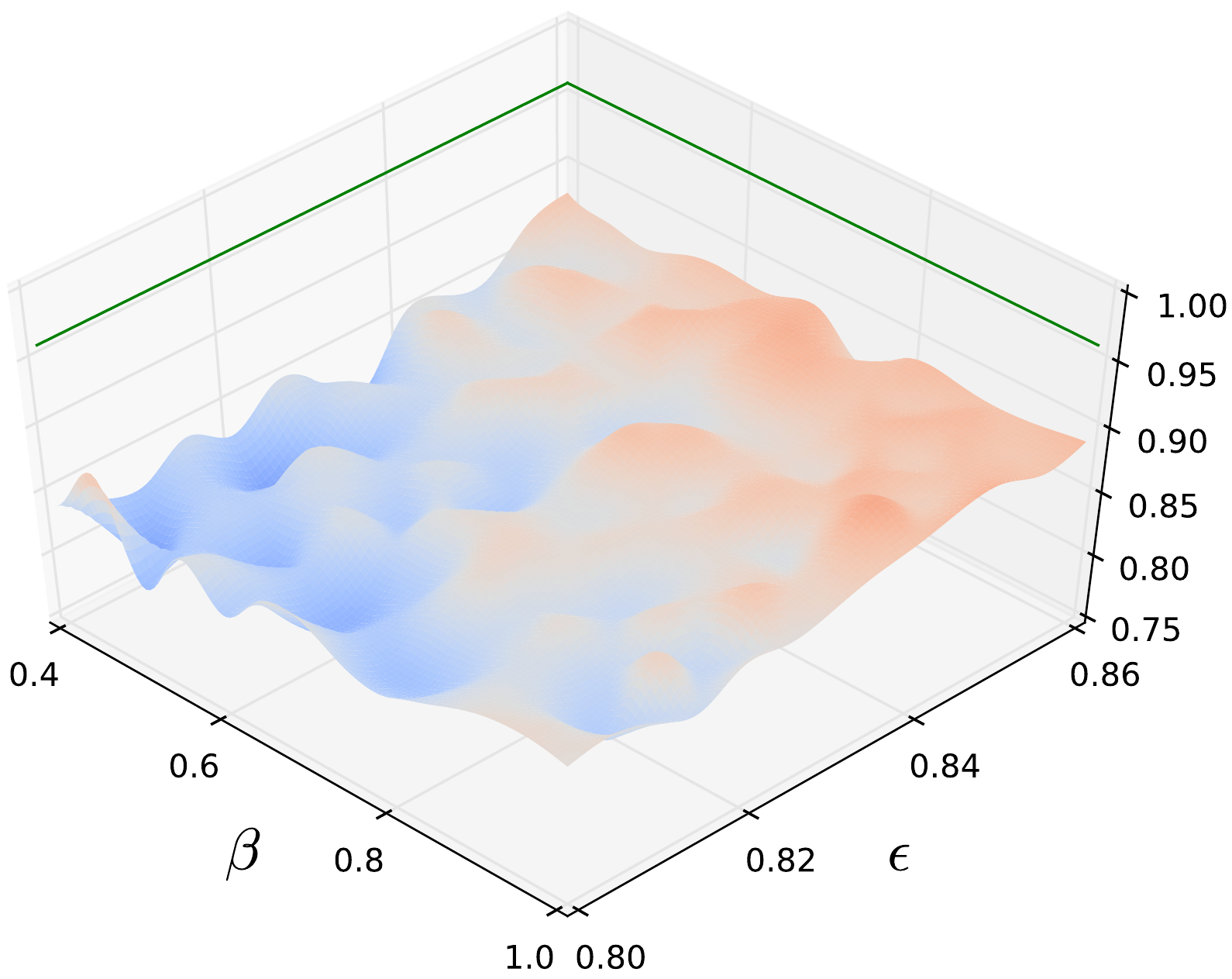}}
\vspace*{-0.05in}
\cap{fig:bang_cifar10s}{BANG CIFAR-10 Models}{These plots summarize our results on CIFAR-10 models trained with BANG using combinations of $\beta$ and $\epsilon$. We tested a grid of those two parameters where $\beta\in\left[0.4,1.0\right]$ with step size $0.05$, and $\epsilon\in\left[0.80,0.86\right]$ with step size $0.005$. We trained a single model with each combination and show \subref*{fig:bang_cifar10s:a} the obtained accuracy on the CIFAR-10 test set, \subref*{fig:bang_cifar10s:b} the achieved success rates by FGS, and the \subref*{fig:bang_cifar10s:c} mean PASS score of HC1 adversarial examples on the CIFAR-10 test images. Each solid green line represents the level of regularly trained learning models. For better visual representation we applied interpolation.}
\end{center}
\vspace*{-0.19in}
\end{figure*}

We also evaluated training with BANG on the so-called ``CIFAR-10 quick'' model of Caffe trained on the CIFAR-10 dataset. CIFAR-10 consists of 60k images, 50k training images, and 10k images used for both validation and testing purposes. The network architecture originally has five layers (three convolutional and two fully connected) that we extended with one Dropout layer, and the learning model is trained with a batch size of 100 for 20k iterations (40 epochs). We use a fixed learning rate of 0.001 that we decrease by a factor of 10 after 36 epochs, and once again after another 2 epochs.

Due to the different nature of CIFAR-10 training, we slightly adjusted BANG parameters.
Specifically, as the classification performance is significantly worse than achieved by LeNet on MNIST yielding proportionately more incorrectly classified samples in each mini-batch, we applied lower local learning rates ($\beta$) and higher values for scaling ($\epsilon$).
Furthermore, we found that scaling incorrectly classified inputs less than correct ones has beneficial effects on robustness, hence, we applied 50\% of the specified $\epsilon$ values on the incorrectly classified batch elements.
Similarly to our conducted experiments on LeNet, we trained classifiers on CIFAR-10 with all possible combinations of $\beta$ and $\epsilon$ parameters of a grid and then measured the accuracy and adversarial vulnerability of each of those networks.
The results are visualized in Figure~\ref{fig:bang_cifar10s}, and for some models obtained with regular training (R0-R1) and optimized with BANG training (B0-B5), we show accuracies and metrics indicating adversarial robustness in Table~\ref{tab:cifar10s}.

As we can see in Table~\ref{tab:cifar10s}, FGS success rates achieved by regular training are significantly decreased by BANG: the rate drops from approximately 96\% to 34\% where, again, the majority of the failed adversarial example generation attempts are due to blank gradients. Figure~\subref{fig:bang_cifar10s:b} shows that as we increase $\epsilon$, the classifiers become more resistant to gradient-based adversarial generation methods. The higher levels of success rates in comparison to LeNet might simply be due to the fact that the classifiers trained on CIFAR-10 are less accurate, therefore, learning the incorrect samples of the batch still has a large contribution on weight updates. While the success rates achieved by HC1 remain high, the quality of HC1 adversarial examples degrades significantly compared to regular training. This degradation is highlighted by both decreasing PASS scores shown in Figure~\subref{fig:bang_cifar10s:c} and by the significantly increased $L_\infty$ norms of adversarial perturbations listed in Table~\ref{tab:cifar10s}. Finally, as shown in Table~\ref{tab:cifar10s}, we can train classifiers with BANG that slightly outperform models of regular training in terms of classification accuracy. Of course, the achieved overall performance depends on the chosen parameters as depicted in Figure~\subref{fig:bang_cifar10s:a}.

\begin{figure*}
\begin{center}
\vspace*{-0.06in}
\centering\subfloat[][\label{fig:cifar10s_noise:a}\centering Regular Training]{\includegraphics[width=0.32\linewidth]{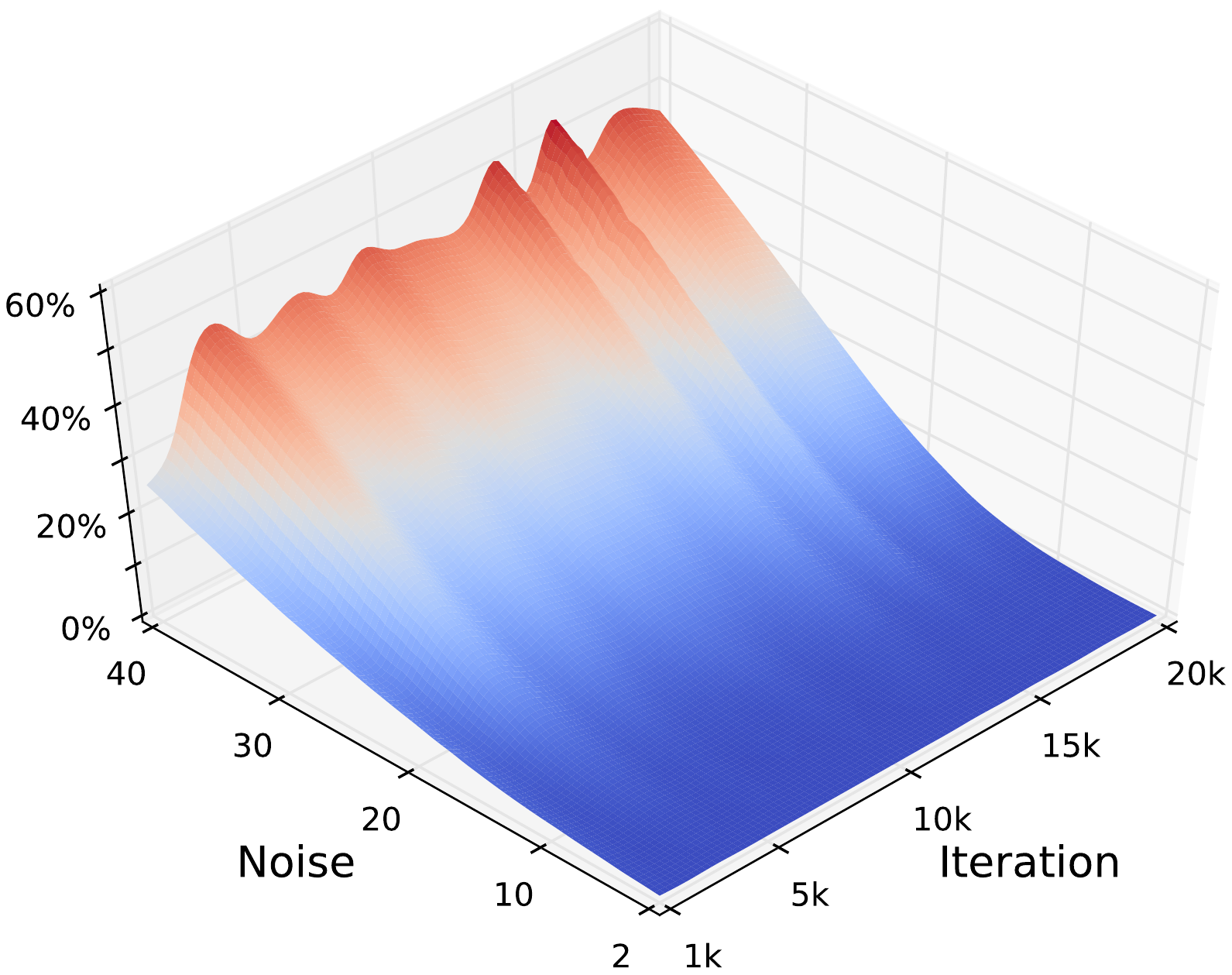}}
\hfill
\centering\subfloat[][\label{fig:cifar10s_noise:b}\centering BANG Training]{\includegraphics[width=0.32\linewidth]{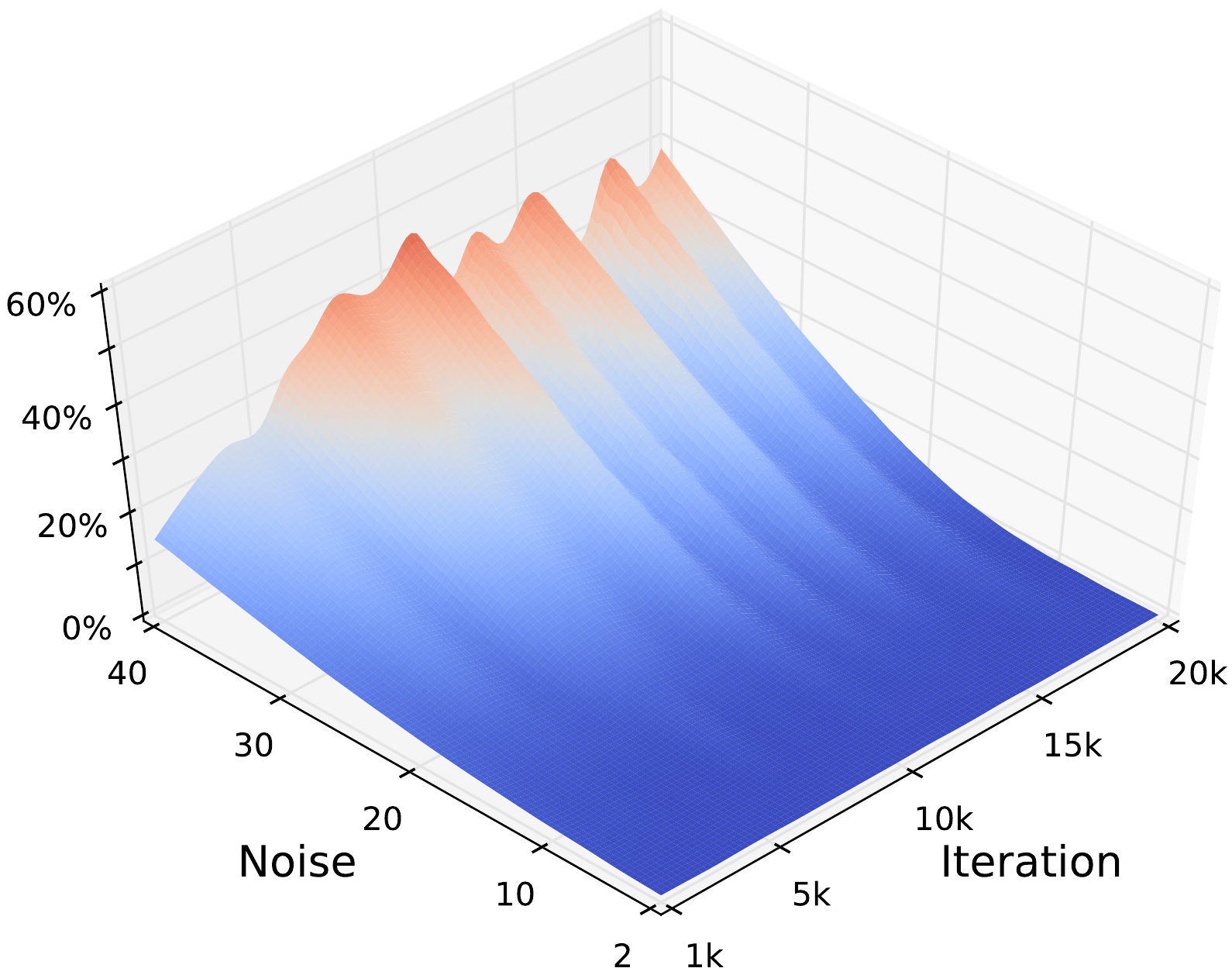}}
\hfill
\centering\subfloat[][\label{fig:cifar10s_noise:c}\centering Absolute Improvement]{\includegraphics[width=0.32\linewidth]{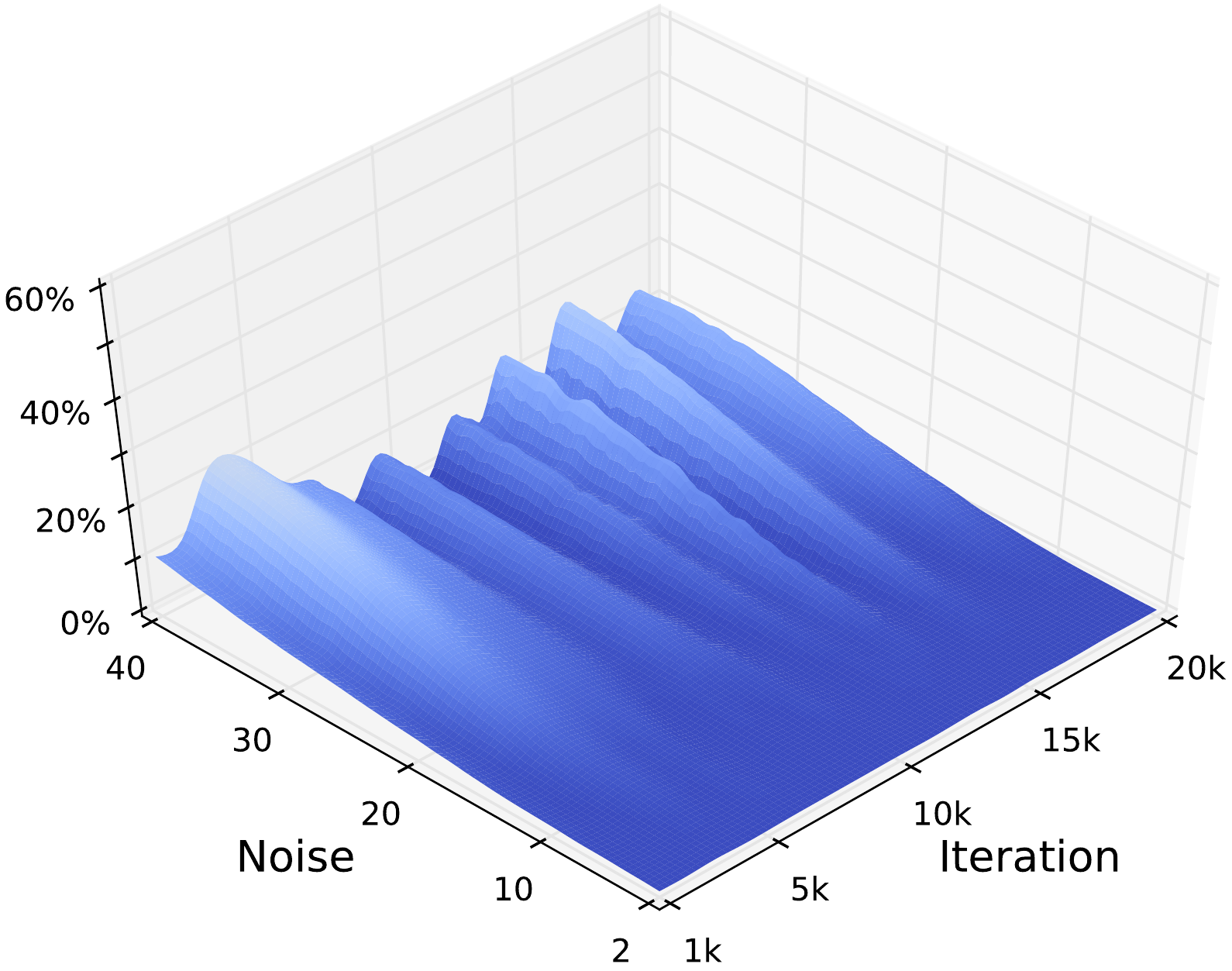}}
\vspace*{-0.05in}
\cap{fig:cifar10s_noise}{CIFAR-10: Robustness to Random Distortions}{These plots show the evolving robustness of CIFAR-10 models: \subref*{fig:cifar10s_noise:a} obtained with regular training (R0 from Table~\ref{tab:cifar10s}), \subref*{fig:cifar10s_noise:b} trained with BANG (B0 from Table~\ref{tab:cifar10s}), and \subref*{fig:cifar10s_noise:c} displays the improvement. After identifying 100 test images per class that are correctly classified by both networks at every second epoch, we perturb each 1000 times with the level of Gaussian noise specified by the standard deviation, and test the networks at different stages of training. The plots show the percentage of distortions yielding misclassifications. For better visual representation we applied interpolation.}
\end{center}
\vspace*{-0.05in}
\end{figure*}

Finally, we ran experiments to better quantify and compare how the robustness of the trained classifiers to random perturbations evolves during training. Similarly to our experiments on LeNet, we selected two classifiers from Table~\ref{tab:cifar10s} for testing: R0 trained regularly and B0 optimized with BANG. We can see in Figure~\subref{fig:cifar10s_noise:a} that the regularly trained R0 model is highly susceptible to larger distortions, its robustness does not improve during training, and finally achieves 46.0\% with respect to the strongest class of Gaussian noise that we formed by using standard deviation of 40 pixels. Contrarily, the B0 model trained with BANG remains more robust throughout training epochs as shown in Figure~\subref{fig:cifar10s_noise:b} and at the end 39.1\% of the strongest distortions change the original classification. The absolute improvements are visualized in Figure~\subref{fig:cifar10s_noise:c}. We can conclude that although BANG enhanced robustness to random perturbations, the results are less impressive in comparison to LeNet -- at least, with respect to the strongest distortions.

\section{Conclusion}

In this paper, we introduced our theory to explain an intriguing property of machine learning models. Namely, the regular training procedure can prevent samples from forming flatter and broader regions around themselves. This evolutionary stalling yields samples remaining close to decision boundaries and, hence, being susceptible to imperceptibly small perturbations causing misclassifications.
To address this problem, we proposed a novel approach to improve the robustness of Deep Neural Networks (DNNs) by slightly modifying the regular training procedure. Our approach does not require additional training data -- neither adversarial examples nor any sort of data augmentation -- to achieve improved robustness, while the overall performance of the trained network is maintained or even enhanced.

We experimentally demonstrated that optimizing DNNs with our Batch Adjusted Network Gradient (BANG) technique leads to significantly enhanced stability in general. By balancing the contributions of batch elements on forming the weight updates, BANG allows training samples to form flatter, more invariant regions around themselves. The trained classifiers become more robust to random distortions, and as we demonstrated with the gradient-based Fast Gradient Sign (FGS) method and the non-gradient-based hot/cold approach where we targeted the closest scoring class (HC1), they are also less vulnerable to adversarial example generation methods.
To visualize the advancement achieved by BANG training in terms of improved adversarial robustness, in Figure~\ref{fig:teaser} correctly classified MNIST and CIFAR-10 test images are presented along with adversarial examples formed via the HC1 approach on DNNs trained regularly and with BANG.
While BANG helps to mitigate adversarial instability, learning models can maintain or even improve their overall classification performance. Our proposed approach achieves these results with negligible computational overhead over the regular training procedure.

Although we managed to achieve good results on two DNNs trained on different datasets, we found that BANG parameters needed to be adjusted to these problems. To obtain better results, exploring the effect of different parameters on different layers, and changing the contributions of correctly and incorrectly classified batch elements can be considered. Future work will focus on having a better understanding of BANG, enhancing the algorithm to be more self-adaptive, and exploring its application for training DNNs on real-world datasets. While some might argue that a similar balancing effect can be achieved by distillation, Carlini\etal{carlini2016defensive} demonstrated that defensive distillation is not effective to improve adversarial robustness.
The effectiveness of BANG to adversarial perturbations obtained via various adversarial example generation techniques likely varies -- as Kurakin\etal{kurakin2017adversarial} observed for adversarial training -- and further research needs to explore that.

In summary, we can conclude that the adversarial instability of DNNs is closely related to the applied training procedures -- as was claimed by Gu\etal{gu2015towards} -- and there is a huge potential in this research area to further advance the generalization properties of machine learning models and their overall performances as well.

\ifwacvfinal
\subsection*{Acknowledgments}

This research is based upon work funded in part by NSF IIS-1320956 and in part by the Office of the Director of National Intelligence (ODNI), Intelligence Advanced Research Projects Activity (IARPA), via IARPA R\&D Contract No. 2014-14071600012. The views and conclusions contained herein are those of the authors and should not be interpreted as necessarily representing the official policies or endorsements, either expressed or implied, of the ODNI, IARPA, or the U.S. Government. The U.S. Government is authorized to reproduce and distribute reprints for Governmental purposes notwithstanding any copyright annotation thereon.

\fi

{\small
\bibliographystyle{ieee}
\bibliography{paper}
}

\end{document}